\documentclass[letterpaper]{article} 
\usepackage[submission]{aaai23}  
\usepackage{times}  
\usepackage{helvet}  
\usepackage{courier}  
\usepackage[hyphens]{url}  
\usepackage{graphicx} 
\usepackage{natbib}  
\usepackage{caption} 

\usepackage{amsfonts}  
\usepackage{subfigure}
\usepackage{amsmath} 
\usepackage{booktabs}
\usepackage{xcolor}    
\usepackage{makecell}
\usepackage{multirow}

\usepackage{algorithm}
\usepackage{algorithmic}

\usepackage{newfloat}
\usepackage{listings}
\DeclareCaptionStyle{ruled}{labelfont=normalfont,labelsep=colon,strut=off} 
\floatstyle{ruled}
\newfloat{listing}{tb}{lst}{}
\floatname{listing}{Listing}
\title{ViT-LSLA: Vision Transformer with Light Self-Limited-Attention}
\author {
    Zhenzhe Hechen,
    Wei Huang,
    Yixin Zhao\textsuperscript{\rm *}
}
\affiliations {
    College of Computer and Information Science, Southwest University, Chongqing, China\\
    zyx\_cq@swu.edu.cn
}

\begin{document}

\maketitle
\begin{abstract}
Transformers have demonstrated a competitive performance across a wide range of vision tasks, while it is very expensive to compute the global self-attention. Many methods limit the range of attention within a local window to reduce computation complexity. However, their approaches cannot save the number of parameters; meanwhile, the self-attention and inner position bias (inside the softmax function) cause each query to focus on similar and close patches. Consequently, this paper presents a light self-limited-attention (LSLA) consisting of a light self-attention mechanism (LSA) to save the computation cost and the number of parameters, and a self-limited-attention mechanism (SLA) to improve the performance. Firstly, the LSA replaces the K (Key) and V (Value) of self-attention with the X(origin input). Applying it in vision Transformers which have encoder architecture and self-attention mechanism, can simplify the computation. Secondly, the SLA has a positional information module and a limited-attention module. The former contains a dynamic scale and an inner position bias to adjust the distribution of the self-attention scores and enhance the positional information. The latter uses an outer position bias after the softmax function to limit some large values of attention weights. Finally, a hierarchical \textbf{Vi}sion \textbf{T}ransformer with \textbf{L}ight \textbf{s}elf-\textbf{L}imited-\textbf{a}ttention (ViT-LSLA) is presented. The experiments show that ViT-LSLA achieves 71.6\% top-1 accuracy on IP102 (2.4\% absolute improvement of Swin-T); 87.2\% top-1 accuracy on Mini-ImageNet (3.7\% absolute improvement of Swin-T). Furthermore, it greatly reduces FLOPs (3.5GFLOPs vs. 4.5GFLOPs of Swin-T) and parameters (18.9M vs. 27.6M of Swin-T). 
\end{abstract}

\section{Introduction}
The advent of Transformer \cite{vaswani2017attention} has made a profound effect on natural language processing (NLP) \cite{devlin2018bert, radford2018gpt, howard2018universal}. In addition, the Vision Transformer (ViT) \cite{dosovitskiy2021an} has shown a promising performance compared with its CNN counterparts. Inspired by the ViT, several visual Transformers were proposed \cite{touvron2021training, wu2021cvt, yuan2021tokens}. However, it is unsuitable for various vision tasks to adopt the primal full self-attention, which results in expensive computation cost (the computational complexity of self-attention is quadratic to image size). 

To address this issue, on the one hand, a typical way is confining the range of global self-attention to a local region. Swin Transformer \cite{liu2021swin} limited the computation of self-attention to local windows and constructed cross-window connections between two successive blocks. CSwin \cite{ dong2021cswin} and Pale Transformer \cite{wu2021pale} designed cross-shaped windows and Pale-shaped windows respectively. Shuffle Transformer \cite{huang2021shuffle} proposed shuffled windows. Axial-DeepLab \cite{wang2020axial} applied two axial-attention layers consecutively for the height-axis and width-axis, improving both global connection and efficient computation. On the other hand, some recent works were devoted to the linearization of self-attention \cite{choromanski2020rethinking, bello2021lambdanetworks, shen2021efficient}. The CoaT\cite{xu2021co} especially proposed a factorized attention mechanism whose computation complexity is quadratic w.r.t. the channel while linear w.r.t. the image size. These methods reduce the computation cost to some degree; nevertheless, they cannot save the number of parameters. 

Furthermore, many previous works adopted a fixed scale value to deal with the large values of the dot product \cite{vaswani2017attention, liu2021swin, dosovitskiy2021an, dong2021cswin, xu2021co, lee2022mpvit}. The fixed scale can prevent the minimal gradients in the softmax function and adjust the variance of attention scores to 1. Because it cannot help self-attention grasp the positional information, in some works \cite{liu2021swin, bao2020unilmv2, hu2018relation}, an inner relative position bias is used to enhance the ability to capture positional information. However, the values of attention scores are the similarities between each vector to the others, which means that the similar vectors have big attention scores. After adding the inner positional bias to the attention scores, the computation of self-attention can be regarded as a local information enhancement. This makes the queries prone to focusing on the similar and close patches including itself rather than really relevant ones.

This paper proposes a Light Self-Limited-Attention (LSLA) consisting of a light self-attention mechanism (LSA) and a self-limited-attention mechanism (SLA). 

Unlike the encoder-decoder architecture and cross-attention mechanism in the natural language processing (NLP) tasks (e.g. machine translation), the vision Transformers have encoder-only architecture and self-attention mechanism in classification tasks. Moreover, there are two kinds of language input in machine translation, but image classification only needs to deal with the image input. Therefore, the LSA changes self-attention calculation from Q, K, V to Q, X, X, which can significantly reduce the parameters and computation cost of the self-attention layer. It can be beneficial to stack more self-attention blocks, which helps Transformers go deeper and get better performance.

The self-limited-attention mechanism (SLA) proceeds as follows. Firstly, the dynamic scale cooperating with the inner position bias can explicitly indicate the positional information (see blue patches in Figure \ref{fig:cat_scale}). Based on it, the outer position bias can powerfully limit some large values of attention weights (see red patches in Figure \ref{fig:cat_out}), which is helpful to pay attention to the meaningful patches instead of the similar but unimportant ones. These processes can be regarded as a local information integration, which is conducive to retaining the diversity of information for each query patch.

Through the LSLA, this paper designs a hierarchical \textbf{Vi}sion \textbf{T}ransformer (as illustrated in Figure \ref{fig:architecture}) with \textbf{L}ight self-\textbf{L}imited-attention (ViT-LSLA), which achieves a better performance than previous approaches and reduces parameters and computation cost significantly. The ViT-LSLA (18.9M, 3.5GFLOPs) achieves 71.6\% Top-1 classification accuracy (2.4\% and 1.2\% absolute improvement of Swin-T (27.6M, 4.5GFLOPs) and MPViT-S (22.6M, 4.8GFLOPs)) on IP102, and 87.2\% Top-1 classification accuracy on Mini-ImageNet (3.7\% and 1.1\% absolute improvement of Swin-T and MPViT-S). The models in this paper were trained on two Tesla P100 GPUs.

The contributions of this work are summarized as follows:
\begin{itemize}
	\item A light self-attention mechanism (LSA) is provided as a plug-and-play module. The current Transformers can save the number of parameters and FLOPs conveniently without losing the accuracy by applying the LSA in self-attention blocks.
	\item A self-limited-attention mechanism (SLA) is introduced. Based on positional information, an outer position bias is adopted to powerfully limit the large attention weights. Thus, Transformers can capture truly meaningful information rather than which just has high similarity.
	\item Establishing a simple Transformer model with the above components can not only remarkably reduce the computation cost and the number of parameters, but also significantly improve the performance.
\end{itemize}

\section{Related Work}
In NLP tasks, there are various efficient Transformer models \cite{beltagy2020longformer, choromanski2020rethinking, katharopoulos2020transformers, kitaev2020reformer, roy2021efficient}. In the CV field, lots of visual Transformers \cite{dosovitskiy2021an, touvron2021training, wu2021cvt, yuan2021tokens} adopted the original full self-attention. However, because the resolution of images is very high, it is essential to design efficient self-attention mechanisms for vision tasks. To avoid the quadratic computation complexity caused by self-attention, many approaches \cite{wang2021pyramid, zhu2020deformable, liu2021swin, huang2021shuffle} are devoted to improving efficiency while maintaining performance.

\subsection{Local Self-Attention Mechanism}
Long-range dependency is important for NLP tasks, because one word may be associated with another far word. However, two patches, which are far apart in an image, may not be relevant (e.g. sky and ground). Namuk Park et al. \cite{park2022how} introduce that an appropriate inductive bias is more profitable to improve the performance of MSA than long-range dependency. As an extreme example, a local 3$\times$3 receptive field outperforms the global one, because it reduces unnecessary degrees of freedom. Many works \cite{liu2021swin, huang2021shuffle, fang2021msg, wang2020axial, dong2021cswin, yang2021focal} focus on strengthening the local feature extraction of Vision Transformers and reducing the quadratic complexity of self-attention. 

\subsection{Efficient Self-attentions Mechanism}
Several works improve efficiency by constraining the sequence length of key and value. Deformable attention \cite{zhu2020deformable} sampled a part of keys from the full set by a linear layer, hoping to get similar global attention. However, such a downsampling operation may result in information confusion. CoaT \cite{xu2021co} proposed an efficient factorized self-attention, which has the linear computation complexity w.r.t the number of image size and quadratic complexity w.r.t the channels. This modification can save quite a few computation cost compared with the original self-attention (whose computation complexity is quadratic w.r.t the number of image size). An efficient self-attention \cite{bello2021lambdanetworks} introduced a global context modeling, which constructed the relative position embeddings in local context modeling by convolutions. Different from the above works, this paper presents a simple light self-attention mechanism, which discards the Key and Value in MSA. 

\subsection{Positional Encoding}
Because self-attention cannot grasp the positional information, it is essential to apply positional encoding for Transformer models. A direct way is to inject absolute positional encoding (APE) into the input embedding \cite{vaswani2017attention}. Relative positional encoding (RPE) \cite{liu2021swin, bao2020unilmv2, hu2018relation} and convolutional positional encoding (CPE) \cite{xu2021co, chu2021conditional} are also widely used in Transformers. All these methods are applied inside the softmax function to reevaluate the attention scores. However, attention scores indicate the values of the dot product, and two similar vectors have a big dot product. It makes attention focus on the near and similar areas. In this paper, a different way of positional encoding is presented, which can be deployed outside the softmax function to reevaluate the attention weights. It can limit the neighbors' big attention weights and maintain meaningful ones.

\begin{figure*}[htbp]
	\centering
	\subfigure{
		\includegraphics[width=0.8\textwidth]{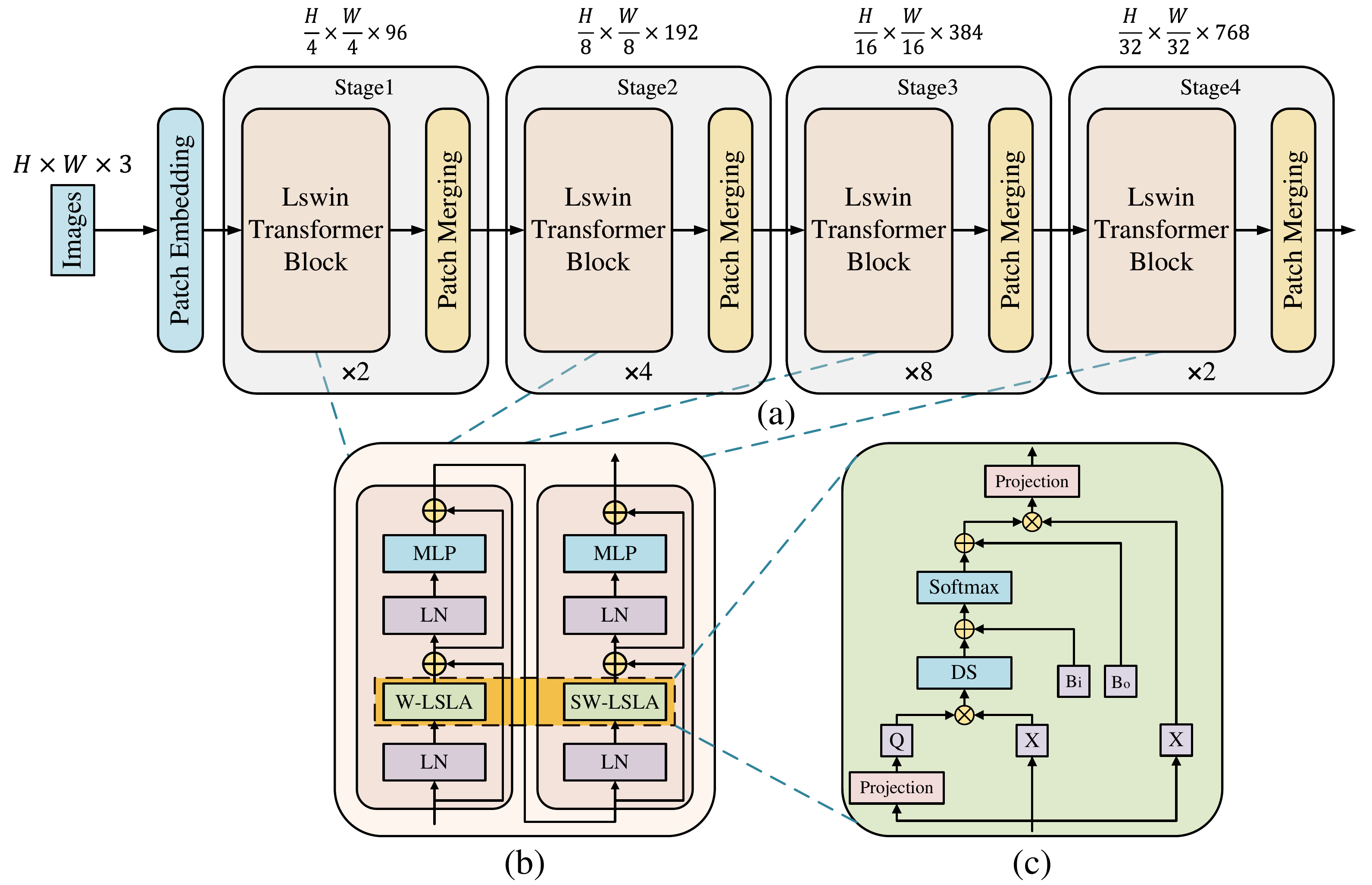}
		\label{fig:architecture}
	}
	\subfigure{
		\label{fig:blocks}
	}
	\subfigure{
		\label{fig:LSLA}
	}
	\caption{(a) is the model architecture of the ViT-LSLA Transformer; (b) shows two successive ViT-LSLA Transformer blocks; (c) is the description of the LSLA. }
\end{figure*}
\section{Method} 
\label{sec:3}

\subsection{Architecture}
The overview architecture of the ViT-LSLA is illustrated in Figure \ref{fig:architecture}. Following \cite{lee2022mpvit}, for an input image with the size of $H \times W \times 3$, a stem block consisting of two $3 \times 3$ convolutional layers is adopted to obtain a feature with the size of $H/4 \times W/4 \times 96$. The whole model is composed of four stages referring to \cite{liu2021swin}. For producing a hierarchical representation, the patch merging layer following \cite{dong2021cswin} is used to reduce the number of tokens and expand the channel dimension between two adjacent stages.

The rough architecture of ViT-LSLA blocks (see Figure \ref{fig:blocks}) follows Swin Transformer blocks. The difference between the blocks of Swin and ViT-LSLA is that the latter replaces the original self-attention mechanism with the light self-limited-attention mechanism (LSLA). 

As shown in Figure \ref{fig:LSLA}, there are the components of our LSLA module; the primary contributions of this paper are the light self-attention mechanism ($QXX$), and the self-limited-attention mechanism consisting of a dynamic scale ($DS$) and an outer position bias ($B_o$). Each of these components is severally elaborated in the following subsections.

\subsection{Light Self-Attention Mechanism}
\label{sec:LSA}
Before introducing the light self-attention (LSA), the FC (fully-connected layers) of the MLP (multilayer perceptron) is firstly considered. It is essential for MLP to apply an activation function after each layer. Otherwise, the MLP will collapse into a linear model \cite{zhang2021dive}:
\begin{equation}
	H = XW_1 + b_1,\\
	\label{eq.hidden}
\end{equation}
\begin{equation}
	O = HW_2 + b_2,\\
	\label{eq.output}
\end{equation}
where $X \in \mathbb{R}^{n\times n}$ is the input; $H$ and $O$ are the hidden variables and the output, and there is no nonlinear activation function between both of them; $W_1 \in \mathbb{R}^{n\times m}$, $W_2 \in \mathbb{R}^{m\times d}$ are the weight matrices, and $b_1 \in \mathbb{R}^{1\times m}$, $b_2 \in \mathbb{R}^{1\times d}$ are the biases. Substituting Eq. (\ref{eq.hidden}) into Eq. (\ref{eq.output}), the following equation can be derived:
\begin{equation}
	\begin{aligned}
		O &= (XW_1+b_1)W_2 + b_2\\
		&= XW_1W_2 + b_1W_2 + b_2.\\
	\end{aligned}
\end{equation}
Adding the hidden layer requires the model to track and update additional sets of parameters, but it is not beneficial to improving performance. Assume that $W = W_1W_2$ and $b=b_1W_2+b_2$. The solution is deduced as:
\begin{equation}
	\begin{aligned}
		O &= XW + b,
	\end{aligned}
\end{equation}
where $W \in \mathbb{R}^{n\times d}$ is the weight matrix and $b \in \mathbb{R}^{1\times d}$ is the bias. It shows that the hidden layer and the output layer can be collapsed into a single layer.

Inspired by this, the original self-attention mechanism can also be simplified in the same way.\\

\noindent\textbf{QX or QK: }This part firstly expounds on how \textbf{K} (Keys) can be easily replaced by \textbf{X}. For the sake of simplicity, the formula of original MSA is defined as below:
\begin{equation}
	\label{eq.qkv}
	\text{Attention}(Q, K, V) = \text{SoftMax}(QK^T)V.
\end{equation}
In Eq. (\ref{eq.qkv}), where $Q$, $K$, and $V$ $\in \mathbb{R}^{M^2\times d}$ respectively are the matrices of Query, Key, and Value; $M^2$ is the number of patches in a window (assume that the $M$ is 7 in this paper) and $d$ is the channel dimension. Moreover, the $Q$, $K$, and $V$ are derived from $X$ via an FC, whose bias is omitted for simplicity:
\begin{equation}
	\label{eq.weight}
	\begin{aligned}
		Q = XW_q,\ K = XW_k,\ V = XW_v,
	\end{aligned}
\end{equation}
where $X \in \mathbb{R}^{M^2\times d}$ is the original input and $W_q, W_k, W_v \in \mathbb{R}^{d\times d}$ are the weight matrices of FC. On the basis of Eq. (\ref{eq.qkv}) and (\ref{eq.weight}), the $QK^T$ can be mathematically formulated as below: 
\begin{equation}
	\begin{aligned}
		QK^T = (QW_k^T)X^T = X(W_qW_k^T)X^T.
	\end{aligned}
	\label{eq.qk}
\end{equation}
Let  $W_qW_k^T$ be $W \in \mathbb{R}^{d\times d}$, the following expression can be derived:
\begin{equation}
	QK^T = XWX^T.
\end{equation}
Let  $XW$ be $\overline{Q} \in \mathbb{R}^{M^2\times d}$, Eq. (\ref{eq.qk}) reduces to:
\begin{equation}
	QK^T = \overline{Q}X^T.
\end{equation}
Note that the $\overline{Q}$ can also be derived from $X$ via an FC; both $Q$ and $\overline{Q}$ can be optimized in the training process. To sum up, the ${\overline{Q}}X^T$ is equivalent to $QK^T$; let $\overline{Q}$ be $Q$, the Eq. (\ref{eq.qkv}) can be rewritten as:
\begin{equation}
	\label{eq.qxv}
	\text{Attention}(Q, X, V) = \text{SoftMax}(QX^T)V.
\end{equation}

\begin{figure}[htbp]
	\centering
	\subfigure[QXX with projection (Ours)]{
		\includegraphics[width=0.2\textwidth]{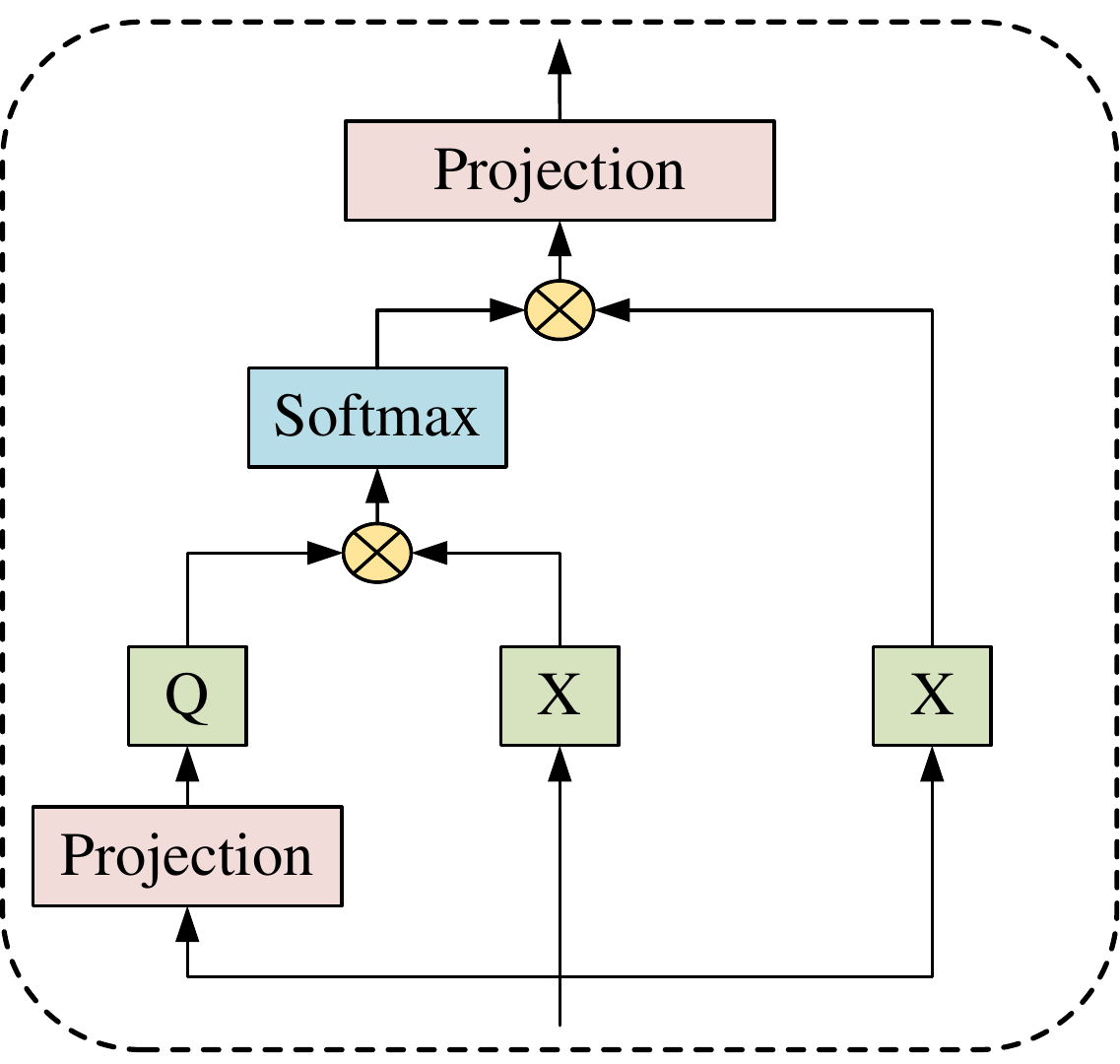}
		\label{fig:qxx}
	}
	\subfigure[QXV without projection]{
		\includegraphics[width=0.2\textwidth]{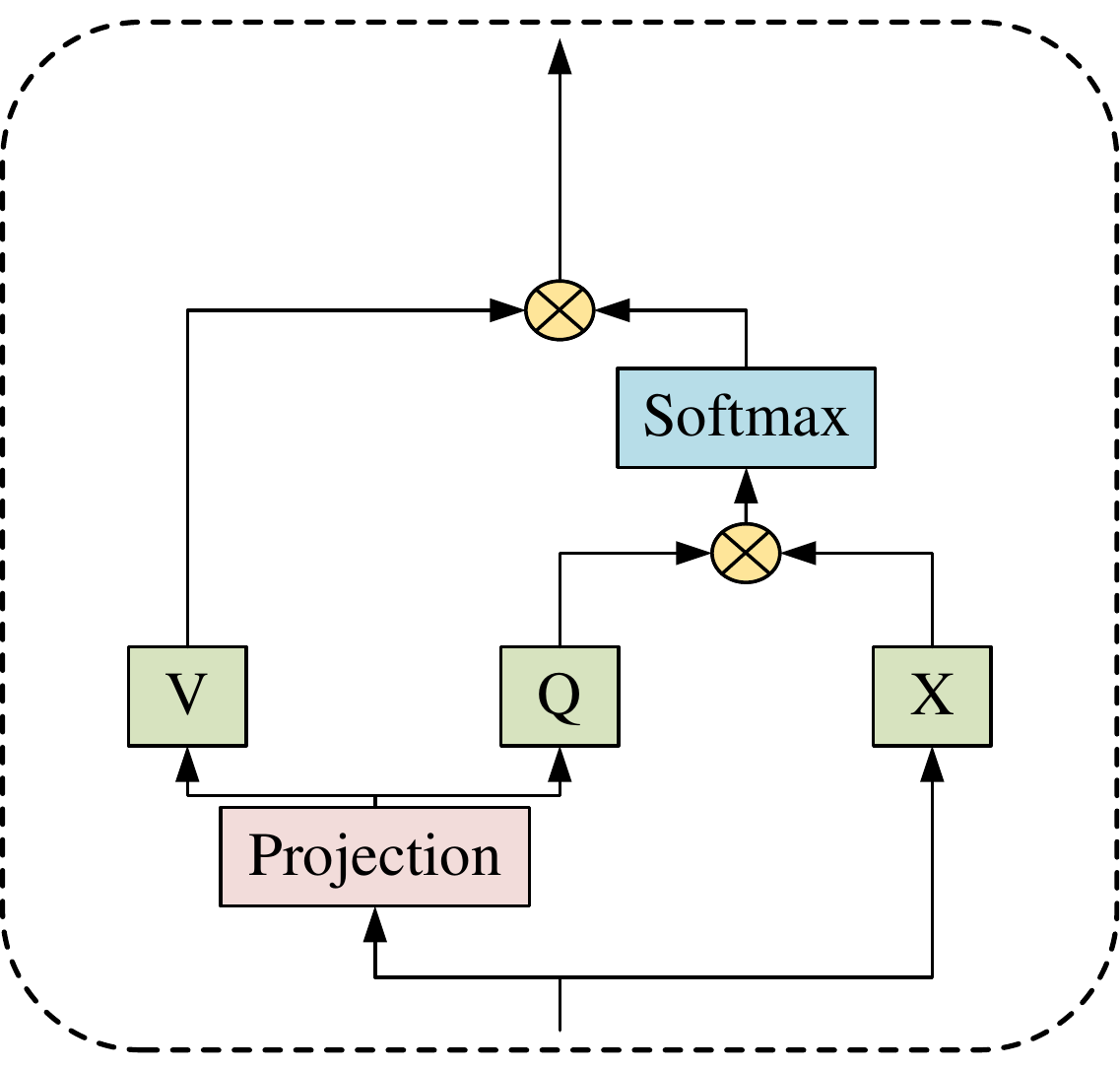}
		\label{fig:qxv}
	}
	\caption{ The rough architecture of QXX and QXV. }
	\label{fig:qxx_qxv}
\end{figure}
\noindent\textbf{QXV or QXX: }Similarly, the QXV can also reduce to QXX in the same way. Eq. (\ref{eq.qxv}) is predigested as:
\begin{equation}
	\text{Attention}(Q, X, V) = AV.
\end{equation}
From Eq. (\ref{eq.weight}), the following equation can be derived: 
\begin{equation}
	AV=AXW_v.
	\label{eq.av}
\end{equation}
It is reasonable for Transformers, which replace $V$ with $X$, to get a weaker performance. However, there is a linear projection applied at the end of the MSA module (see Figure \ref{fig:qxx}), which is the point why can $V$ be replaced with $X$. 

Let the weight matrix of this linear projection be $W_o \in \mathbb{R}^{d\times d}$. The following equation can be derived:
\begin{equation}
	AVW_o = AXW_vW_o.
\end{equation}
Let $W_vW_o$ be $W \in \mathbb{R}^{d\times d}$, it can be expressed as:
\begin{equation}
	AXW_vW_o = AXW.
\end{equation}
Therefore, it is rational to conclude that the linear projection of $V$ ($W_v$) is equivalent to the last one ($W_o$); namely, either of them can be discarded as shown in Figures \ref{fig:qxx} and \ref{fig:qxv}. In this paper, the first method is selected; therefore, the light self-attention (LSA) mechanism is defined as:
\begin{equation}
	\label{eq.qxx}
	\text{Attention}(Q, X, X) = \text{SoftMax}(QX^T)X.
\end{equation}
\subsection{Self-Limited-Attention Mechanism}
\label{sec:3.3}
The Self-Limited-Attention (SLA) has a positional information module and a limited-attention module. The former contains a dynamic scale and an inner position bias to enhance the positional information, which can mark the patches that need to be limited. Based on it, the latter adopts an outer position bias after the softmax function to limit the values of attention weights marked before. \\

\textbf{Positional Information Module: }Since the Transformer models cannot capture the positional information, it is necessary to apply some relative or absolute position for the image patches. In \cite{liu2021swin, bao2020unilmv2, hu2018relation} a relative position bias is introduced to each head in computing similarity. This paper follows \cite{liu2021swin} by maintaining an inner position bias (inside the softmax function). The LSA with fixed scale (FS) is formulated as:
\begin{equation}
	\label{eq.qxx_fs}
	\text{Attention}(Q, X, X) = \text{SoftMax}(QX^T/\sqrt{d}+B_i)X,
\end{equation}
where $\sqrt{d}$ is the query dimension; $B_i \in \mathbb{R}^{M^2\times{M^2}}$ is the inner position bias. 

\begin{figure}[htbp]
	\centering
	\subfigure[The dynamic scale and inner position bias at the eighth query patch]{
		\includegraphics[width=0.2\textwidth]{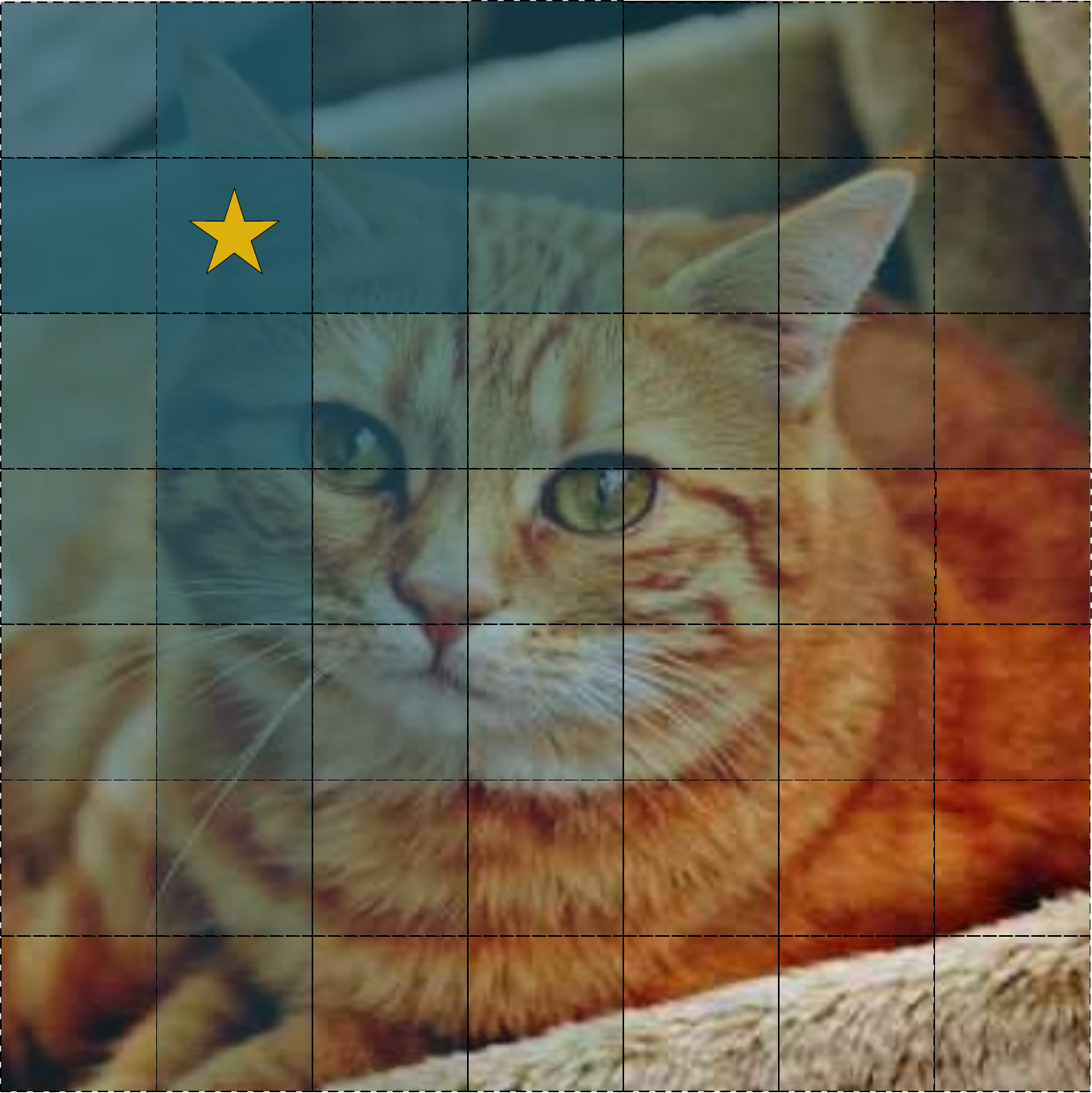}
		\label{fig:cat_scale}
	}
	\hspace{1em}
	\subfigure[The neighbors' attention weights of the query are limited by the outer position bias]{
		\includegraphics[width=0.2\textwidth]{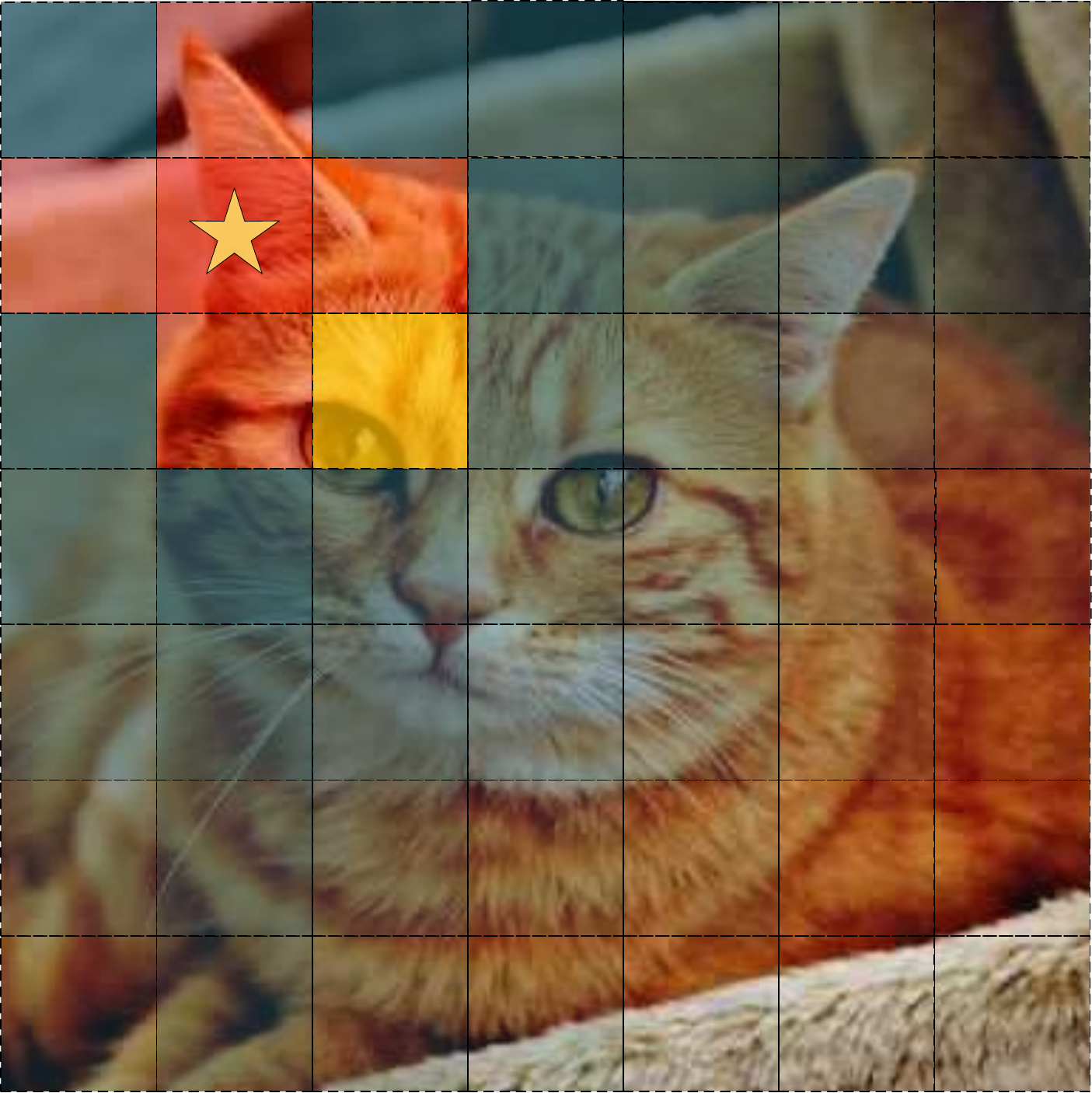}
		\label{fig:cat_out}
	}
	\caption{In(a), it represents the DS and inner position bias enhancing the local positional information; the patches close to the query have stronger DS and inner position bias. (b) indicates the outer position bias can powerfully limit large neighbors' attention weights of queries and retain the truly meaningful ones. }
	\label{fig:cat}
\end{figure}

To enhance the ability to capture positional information, a dynamic scale (DS) is designed to replace the fixed scale. It concerts with the inner position bias, containing a group of learnable parameters. The LSA with DS can be formulated as follows:
\begin{equation}
	\label{eq.qxx_ds}
	\text{Attention}(Q, X, X) = \text{SoftMax}(QX^T\times{DS}+B_i)X,
\end{equation}
where $DS \in \mathbb{R}^{M^2\times{M^2}}$, it denotes that the patches near the query patch (golden star) have larger values of dynamic scale and inner position bias, as shown in Figures \ref{fig:cat_scale} and \ref{fig:scale_value}. It is helpful for the query patches to capture the positional information of each image patch. This can be regarded as a local information enhancement.\\

\textbf{Limited-Attention Module: } Depending on the positional information module, LSA can enhance the local information. Nevertheless, the values of dot products are the similarities between each vector to the others, which means that the similar vectors have big attention weights. However, such a local information enhancement forces self-attention to focus on the close and similar patches, including itself.

To retain the diversity of information for each query patch, an additional outer position bias $B_o \in \mathbb{R}^{M^2\times{M^2}}$ (outside the softmax function) is used, whose construction is as same as the inner one. The outer position bias cooperating with the dynamic scale and inner position bias can overwhelmingly restrict the big attention weights close to the query patch. As shown in Figures \ref{fig:cat_out} and \ref{fig:attn_weight}, the eighth patch is the query. The outer position bias can severely limit its neighbors' attention weights, especially its own. But the attention weights of the other patches are retained. Finally, the light self-limited-attention (LSLA) is formulated as below:
\begin{equation}
	\label{eq:final_qxx}
	\text{Attention}(Q, X, X) = (\text{SoftMax}(QX^T\times{DS}+B_i)+B_o)X.
\end{equation}
It is easy to understand the rationality of outer position bias. In Figure \ref{fig:cat_out}, the query patch (golden star) locates at the cat's ear and the notable patch (golden patch) locates at the cat's eye. This process can be regarded as a local information integration, which is conducive to self-attention to capture the notable information instead of a close and similar one. The code of the LSLA is shown in Algorithm \ref{alg:code}.
\begin{algorithm}[!ht]
	\caption{Light Self-Limited-Attention}
	\label{alg:code}
	\definecolor{codeblue}{rgb}{0.25,0.5,0.5}
	\lstset{
		basicstyle={\footnotesize\ttfamily},
		numbers=left,numberstyle=\footnotesize,xleftmargin=2em,
		columns=fullflexible,
		aboveskip=0pt,belowskip=0pt,
		showstringspaces=false,tabsize=2,breaklines=true,
		morecomment=[l][\color{codeblue}]{\%},
		commentstyle=\color{codeblue},
		morekeywords={def, return},
		keywordstyle=\color[rgb]{0.8,0.2,0.2},
	}
	\begin{lstlisting}[]
% Input: The feature map of image, shape [N, D]: x
% Output: The result of light self-attention on input, shape [N, D]: y
% Signal meanings:
	% N: the number of input tokens
	% H: the number of attention heads
	% HD: the number of head-dimension
	% D: the dimension of token vector
	% @: matrix multiplication
def init():
	fc_q = Linear(D, D)
	inner_bias = Parameter(size=(H, N, N))
	outer_bias = Parameter(size=(H, N, N))
	dynamic_scale = Parameter(size=(N, N))
	proj = Linear(D, D)
	
def forward(x):
% First step: compute attention scores
	% linearly map the tokens to query
	q = fc_q(x)  % shape: [N, D]
	% split into multi-head
	q = q.reshape(H, N, HD)
	x = x.reshape(H, N, HD)
	% compute the attention scores
	% shape: [H, N, N]
	attn = (q @ x.transpose())
% Second step: capture positional information
	attn = attn * dynamic_scale 
	attn = attn + inner_bias
% Third step: compute attention weights
	attn = softmax(attn, dim=-1)
	% Limit neighbours' attention weights by outer position bias
	attn = attn + outer_bias
	% Final step: concatenate all heads
	y = (attn @ x).concat()  % shape: [N, D]
	y = proj(y)
	return y
	\end{lstlisting}
\end{algorithm}

\section{Experiments}
To show the effectiveness of ViT-LSLA, several experiments are conducted on different image classification datasets, and the training settings are introduced. Furthermore, detailed ablation studies are performed to analyze every module of ViT-LSLA.

\subsection{Image Classification}
\noindent\textbf{Datasets: }This paper evaluates ViT-LSLA on three benchmark datasets, IP102 \cite{wu2019ip102}, Food101 \cite{bossard14}, and Mini-ImageNet \cite{ravi2016optimization}.\\

\noindent\textbf{Experiments Setup: }Both training and evaluation are conducted with the input size of $224\times224$ on all datasets above. All models train 300 epochs on the three datasets; the training setting follows \cite{liu2021swin}. The AdamW optimizer with a weight decay of 0.05 is used. The default batch size and initial learning rate are 256 and 2.5e-4; except that 192 and 1.875e-4 of MPViT-S (due to the limit of graphic memory). All experiments are performed with two Tesla P100 GPUs.\\

\noindent\textbf{IP102: }IP102 is a large-scale dataset of insect pets, which consists of more than 75,000 images on 102 insect pets. For classification, it has 45,095 images for training and 22619 testing images. As illustrated in Table \ref{tab:IP102}, this paper compares some recent SOTA Vision Transformers with our ViT-LSLA. The ViT-LSLA outperforms all of the other SOTA counterparts. The ViT-LSLA is +2.4\% higher than the most relevant Swin-T Transformer in top-1 accuracy; +2.6\%, 3.4\%, and 1.2\% better than CSwin-T, CoarT-Lite-Small, MPViT-S respectively. Furthermore, the ViT-LSLA can save the number of parameters and computation cost significantly; 18.9M (-8.7M) of ViT-LSLA vs. 27.6M of Swin-T, which has almost one-third fewer parameters. Meanwhile, among these models, ViT-LSLA has the fewest FLOPs, which is almost three-quarters of Swin-T. To summarize, the ViT-LSLA achieves a better efficiency-quality trade-off compared with other SOTA Vision Transformers.

\begin{table}[htbp]
	\centering
	\small
	\begin{tabular}{c|ccc|c}
		\toprule
		\multicolumn{5}{c}{\textbf{IP102 trained models}} \\
		\hline
		Model & \begin{tabular}[c]{@{}c@{}}Input\end{tabular} & Params & FLOPs &  \begin{tabular}[c]{@{}c@{}} Top-1 acc \end{tabular} \\
		\hline
		Swin-T\shortcite{liu2021swin} & 224$^2$ & 27.6M & 4.5G & 69.2 \\
		CSwin-T\shortcite{dong2021cswin} & 224$^2$ & 21.9M & 4.3G & 69.0 \\
		CoaT-Lite-S\shortcite{xu2021co} & 224$^2$ & 19.4M & 4.0G & 68.2 \\
		MPViT-S\shortcite{lee2022mpvit} & 224$^2$ & 22.6M & 4.8G & 70.4 \\
		ViT-LSLA (Ours) & 224$^2$ & 18.9M & 3.5G & \textbf{71.6} \\
		\hline
		\Xhline{1.0pt}
	\end{tabular}
	\caption{The performance comparison of different backbones on IP102 classification.}
	\label{tab:IP102}
	\normalsize
\end{table}

\noindent\textbf{Food101: }This dataset contains 101,000 real-world food images for 101 of the most popular dishes. There are 750 training images and 250 test images for each class. Table \ref{tab:Food101} demonstrates that the performance of ViT-LSLA is still the most competitive compared to other Transformer counterparts. Because of the marginal utility, the accuracy of ViT-LSLA is not much better than others, but it achieves the most promising efficiency-quality trade-off.

\begin{table}[htbp]
	\centering
	\small
	\begin{tabular}{c|ccc|c}
		\toprule
		\multicolumn{5}{c}{\textbf{Food101 trained models}} \\
		\hline
		Model & \begin{tabular}[c]{@{}c@{}}Input\end{tabular} & Params & FLOPs &  \begin{tabular}[c]{@{}c@{}} Top-1 acc \end{tabular} \\
		\hline
		Swin-T\shortcite{liu2021swin} & 224$^2$ & 27.6M & 4.5G & 88.3 \\
		CSwin-T\shortcite{dong2021cswin} & 224$^2$ & 21.9M & 4.3G & 89.1 \\
		CoaT-Lite-S\shortcite{xu2021co} & 224$^2$ & 19.4M & 4.0G & 85.1 \\
		MPViT-S\shortcite{lee2022mpvit} & 224$^2$ & 22.6M & 4.8G & 88.3 \\
		ViT-LSLA (Ours) & 224$^2$ & 18.9M & 3.5G & \textbf{89.4} \\
		\hline
		\Xhline{1.0pt}
	\end{tabular}
	\caption{The performance comparison of different backbones on Food101 classification.}
	\label{tab:Food101}
	\normalsize
\end{table}

\noindent\textbf{Mini-ImageNet: }This dataset with 100 classes used in few-shot learning research is a subset of ImageNet, which has 600 examples for each category. In this paper, a preprocessed version of miniImageNet in which images are not resized to any particular size is used in supervised learning research; there are 480 images for training and 120 images for tests in each class. As shown in Table \ref{tab:mini-imagenet}, the ViT-LSLA can not only obtain competitive performance on fine-grained classification datasets of IP102 and Food101 but also on generic one of Mini-ImageNet.
\begin{table}[htbp]
	\centering
	\small
	\begin{tabular}{c|ccc|c}
		\toprule
		\multicolumn{5}{c}{\textbf{Mini-ImageNet trained models}} \\
		\hline
		Model & \begin{tabular}[c]{@{}c@{}}Input\end{tabular} & Params & FLOPs &  \begin{tabular}[c]{@{}c@{}} Top-1 acc \end{tabular} \\
		\hline
		Swin-T\shortcite{liu2021swin} & 224$^2$ & 27.6M & 4.5G & 83.5 \\
		CSwin-T\shortcite{dong2021cswin} & 224$^2$ & 21.9M & 4.3G & 84.4 \\
		CoaT-Lite-S\shortcite{xu2021co} & 224$^2$ & 19.4M & 4.0G & 83.7 \\
		MPViT-S\shortcite{lee2022mpvit} & 224$^2$ & 22.6M & 4.8G & 86.1 \\
		ViT-LSLA (Ours) & 224$^2$ & 18.9M & 3.5G & \textbf{87.2} \\
		\hline
		\Xhline{1.0pt}
	\end{tabular}
	\caption{The performance comparison of different backbones on Mini-ImageNet classification.}
	\label{tab:mini-imagenet}
	\normalsize
\end{table}

\subsection{Ablation Study}
In this subsection, the ablation studies are designed to prove the effectiveness of each component in the ViT-LSLA using the IP102 and Mini-ImageNet classification datasets. They are respectively fine-grained and generic classification datasets. It shows that the important elements of LSLA can deal with the different kinds of datasets.\\

\noindent\textbf{Light Self-Attention Mechanism: }In subsection \ref{sec:LSA}, the effect of LSA has been demonstrated by mathematical techniques. As shown in Table \ref{tab:qxx_vs_qkv}, there are several comparisons between LSA and original MSA on the ViT-LSLA. \\

\textbf{QXX or QKV: }Comparing the QXX with QKV, the LSA can significantly save the number of parameters and computation cost by large margins of over 20\% (18.9M vs. 24.0M on Parameters and 3.5GFLOPs vs. 4.4GFLOPs on computation); meanwhile, the accuracy is almost unchanged. The QXX has the same accuracy as QKV on IP102 and a -0.5\% decrease on Mini-ImageNet. Through this control experiment, the result shows the LSA achieves a very promising efficiency-quality trade-off.

\begin{table}[htbp]
	\centering
	\small
	\begin{tabular}{c|cccc|ccc}
	\toprule
	\multicolumn{8}{c}{\textbf{Comparison of different self-attention}} \\
	\hline
	\multirow{3}{*}{\shortstack{Datasets}} & \begin{tabular}[c]{@{}c@{}} \multirow{3}{*}{\shortstack{Q\\X\\X}} \end{tabular} & \multirow{3}{*}{\shortstack{Q\\X\\V}} & \multirow{3}{*}{\shortstack{Q\\K\\V}} & \multirow{3}{*}{NP} &  \begin{tabular}[c]{@{}c@{}} \multirow{3}{*}{\shortstack{Params\\(M)}} \end{tabular} & \multirow{3}{*}{\shortstack{FLOPs\\(G)}} & \multirow{3}{*}{\shortstack{Top-1\\acc}} \\[18pt]
	\hline
	\multirow{4}{*}{IP102}
				& \checkmark & & & \checkmark & 16.4 & 3.1 & 69.8 \\
                & \checkmark & & & & 18.9 & 3.5 & \textbf{71.6} \\
                & & \checkmark & & & 18.9 & 3.5 & 71.3 \\
                & & & \checkmark & & 24.0 & 4.4 & 71.6 \\
	\hline
	\multirow{4}{*}{\shortstack{Mini-\\Image\\Net}}
			   & \checkmark & & & \checkmark & 16.4 & 3.1 & 84.8 \\
               & \checkmark & & & & 18.9 & 3.5 & \textbf{87.2} \\
			   & & \checkmark & & & 18.9 & 3.5 & 87.2 \\
			   & & & \checkmark & & 24.0 & 4.4 & 87.7 \\
	\hline
	\Xhline{1.0pt}
	\end{tabular}
	\caption{This Table shows the performance of QXX, QXV, and QKV on IP102 and Mini-ImageNet classification. NP means the last projection (as shown in Figure \ref{fig:qxx}) of self-attention is discarded.}
	\label{tab:qxx_vs_qkv}
	\normalsize
\end{table}
\textbf{QXX or QXV: }As demonstrated in Table \ref{tab:qxx_vs_qkv}, the ViT-LSLA with QXX has the same number of parameters and computation cost as that with QXV; meanwhile, their performances are almost identical. But the QXX without the last projection will lose accuracy substantially (-1.8\% on IP102 and -2.4\% on Mini-ImageNet). Therefore, the result proves that the QXX with the last projection is equivalent to the QXV (see Figure \ref{fig:qxx_qxv}). 

Since the better performance of the parallel process in QXV, it can save memory and training time over QXX. But on IP102, the QXV has a slightly inferior accuracy compared to the QXX. Thus, in this paper, the QXX is selected as LSA; research on QXV self-attention mechanism is left for future work.\\
\begin{figure}[htbp]
	\centering
	\subfigure[Please see this figure combining with Figure \ref{fig:cat_scale}]{
		\includegraphics[width=0.21\textwidth]{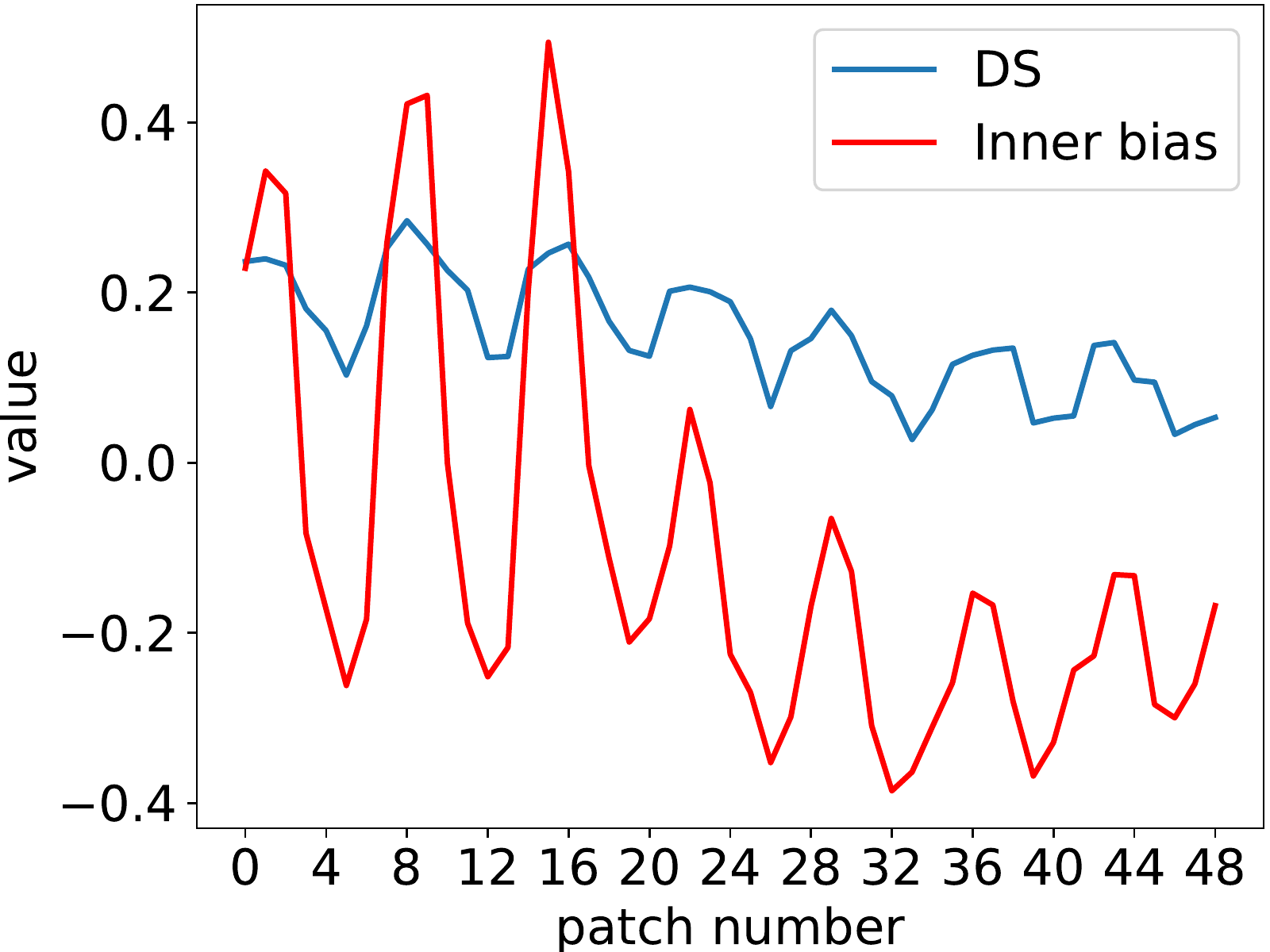}
		\label{fig:scale_value}
	}
	\hspace{0.3em}
	\subfigure[Please see this figure combining with Figure \ref{fig:cat_out}]{
		\includegraphics[width=0.21\textwidth]{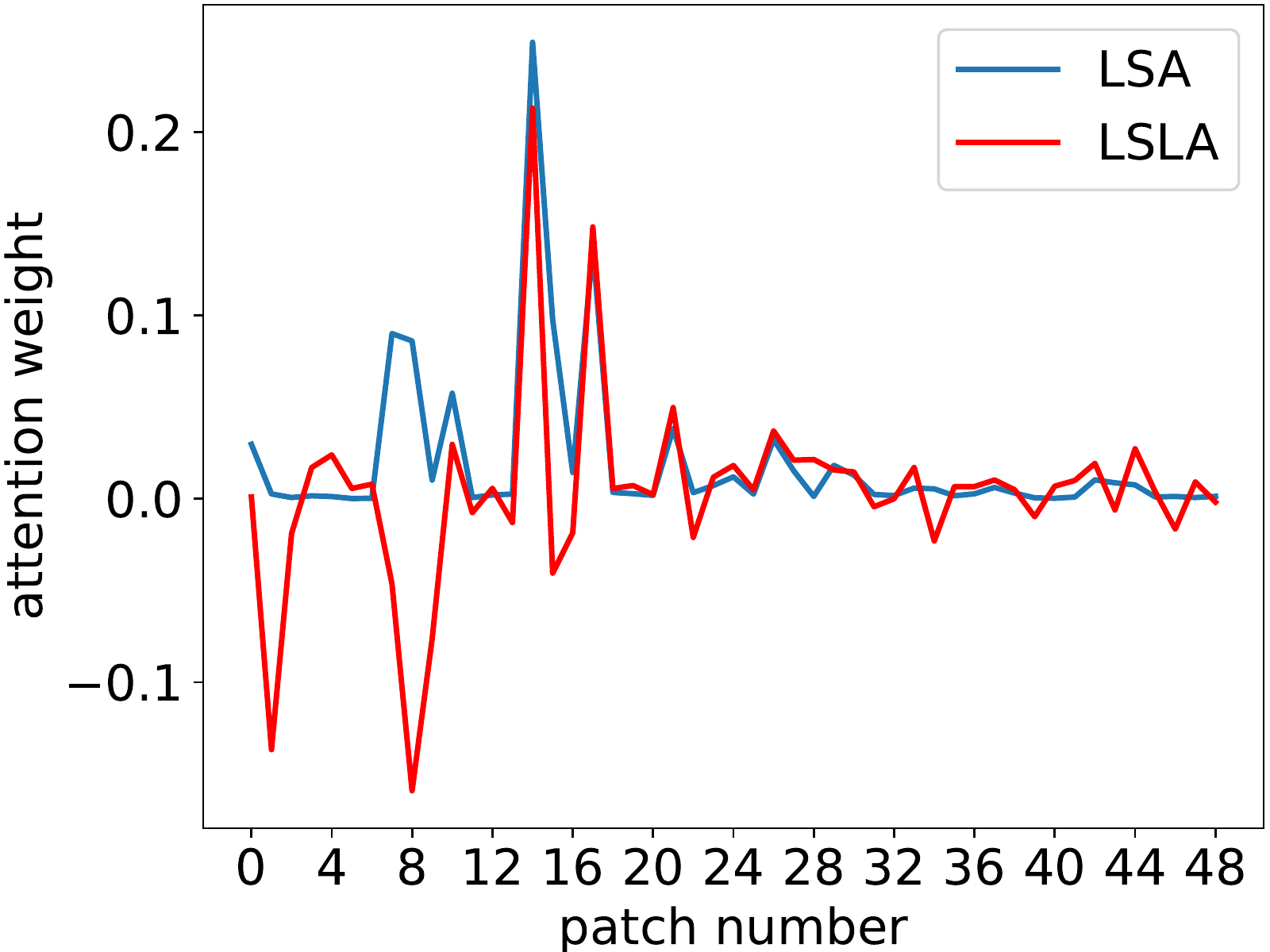}
		\label{fig:attn_weight}
	}
	\caption{ The visualization of the Figure \ref{fig:cat}. The data of Figures (a) and (b) comes from a local window (window size is 7) of ViT-LSLA trained on IP102. }
\end{figure}

\noindent\textbf{Self-Limited-Attention Mechanism: }The SLA consists of a positional information module and a limited-attention module. The former contains a dynamic scale and an inner position bias to enhance the positional information, which can mark the patches that need to be limited. The latter adopts an outer position bias after the softmax function to limit the values of attention weights marked before. \\

\textbf{Positional Information Module: }As illustrated in Figure \ref{fig:scale_value}, the blue line and red line mean that patches close to the query patch have larger values of scale and inner position bias, which infers that the DS and inner position bias contains positional information.\\

\textbf{Limited-Attention Module: }The blue line and red line are severally the attention weights of LSA and LSLA (LSA has no outer position bias compared with LSLA). As shown in Figure \ref{fig:attn_weight}, the outer position bias can powerfully limit neighbors' large attention weights and retain the truly meaningful ones.

\begin{table}[!htbp]
	\centering
	\small
	\begin{tabular}{c|cccc|c}
		\toprule
		\multicolumn{6}{c}{\textbf{Ablation study of ViT-LSLA on IP102 and Mini-ImageNet}} \\
		\hline
		Dataset & \begin{tabular}[c]{@{}c@{}}FS\end{tabular} & DS & inner bias & outer bias & \begin{tabular}[c]{@{}c@{}} Top-1 acc \end{tabular} \\
		\hline
		\multirow{6}{*}{IP102} & \checkmark & & \checkmark & & 70.6 \\
		& \checkmark & & & \checkmark & 70.9 \\
		& \checkmark & & \checkmark & \checkmark & 71.5 \\
		& & \checkmark & \checkmark & & 70.5 \\
		& & \checkmark & & \checkmark & 71.6 \\
		& & \checkmark & \checkmark & \checkmark & \textbf{71.6} \\
		\hline
		\multirow{6}{*}{\shortstack{Mini-\\ImageNet}} & \checkmark & & \checkmark & & 86.1 \\
		& \checkmark & & & \checkmark & 86.9 \\
		& \checkmark & & \checkmark & \checkmark & 87.0 \\
		& & \checkmark & \checkmark & & 86.2 \\
		& & \checkmark & & \checkmark & 87.1 \\
		& & \checkmark & \checkmark & \checkmark & \textbf{87.2} \\
		\Xhline{1.0pt}
	\end{tabular}
	\caption{ The FS and DS are severally the fixed scale and dynamic scale; the inner bias and outer bias are respectively inner and outer relative position bias. }
	\label{tab:Ablation}
	\normalsize
\end{table}
As shown in Table \ref{tab:Ablation}, there are several control experiments on IP102 and Mini-ImageNet datasets. It can be observed that the outer position bias can significantly improve the performance of ViT-LSLA.

In the past works, the common sense is that injecting positional information into self-attention is important. However, the self-attention makes queries naturally prone to focus on similar patches; after adding to the inner position bias, the queries pay more attention to the near and similar areas. This process is a local information enhancement but may cause a lack of information diversity.

First of all, even though there is not any positional information in the LSA, relying only on the outer position bias can obtain better performance. (see the comparison of inner bias and outer bias with a fixed scale in Table \ref{tab:Ablation})

Secondly, adopting inner position bias or dynamic scale to enhance the positional information can mark the patches that need to be limited. (see Figures \ref{fig:cat_out}, \ref{fig:attn_weight}). Based on it, the outer position bias can precisely limit the values of attention weights marked before, which is conducive to retaining the diversity of information for each query patch.

\begin{figure}[htbp]
	\centering
	\subfigure[The accuracy of different models on IP102]{
		\includegraphics[width=0.21\textwidth]{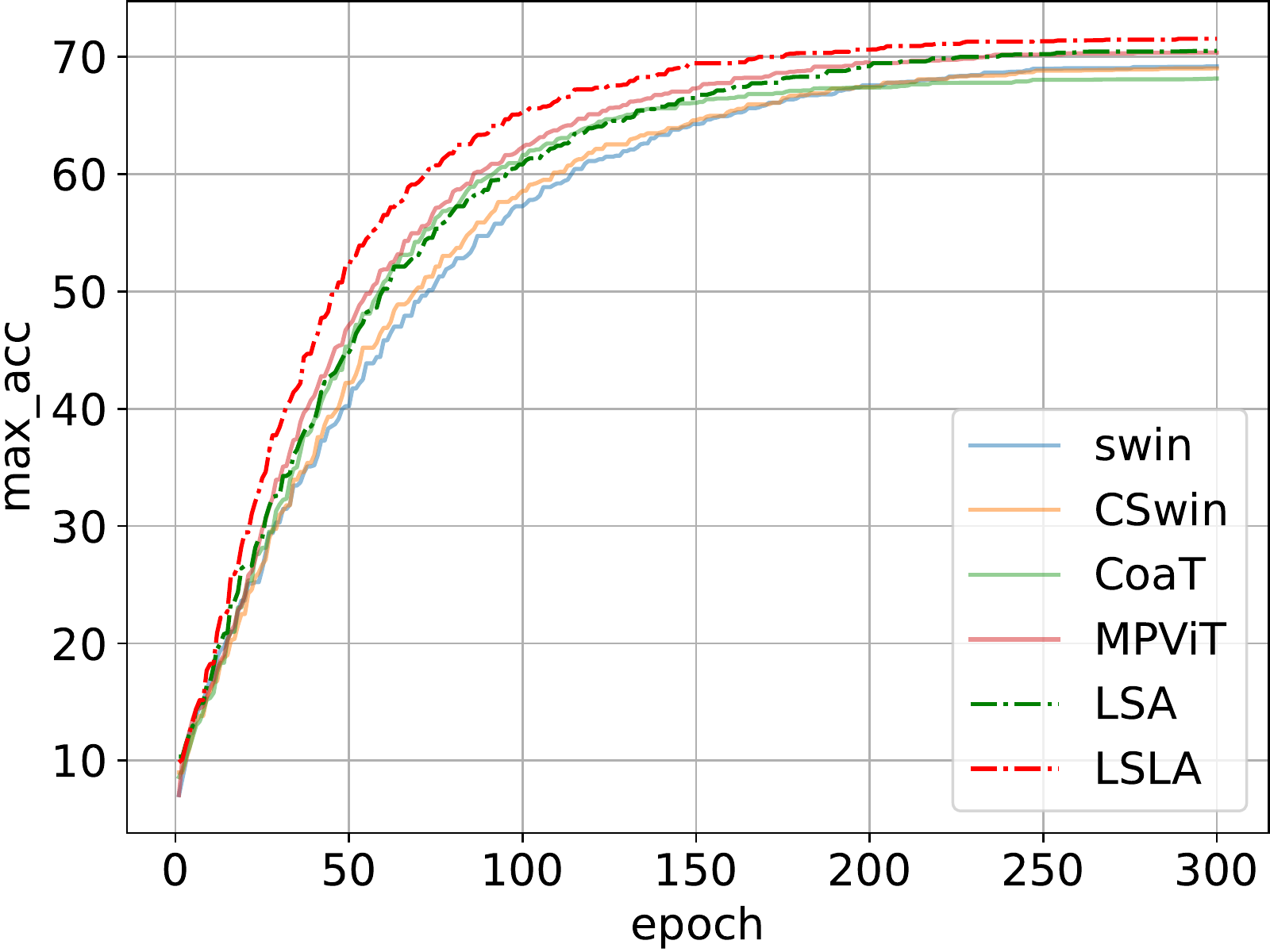}
		\label{fig:train_IP102}
	}
	\hspace{0.5em}
	\subfigure[The accuracy of different models on Mini-ImageNet]{
		\includegraphics[width=0.21\textwidth]{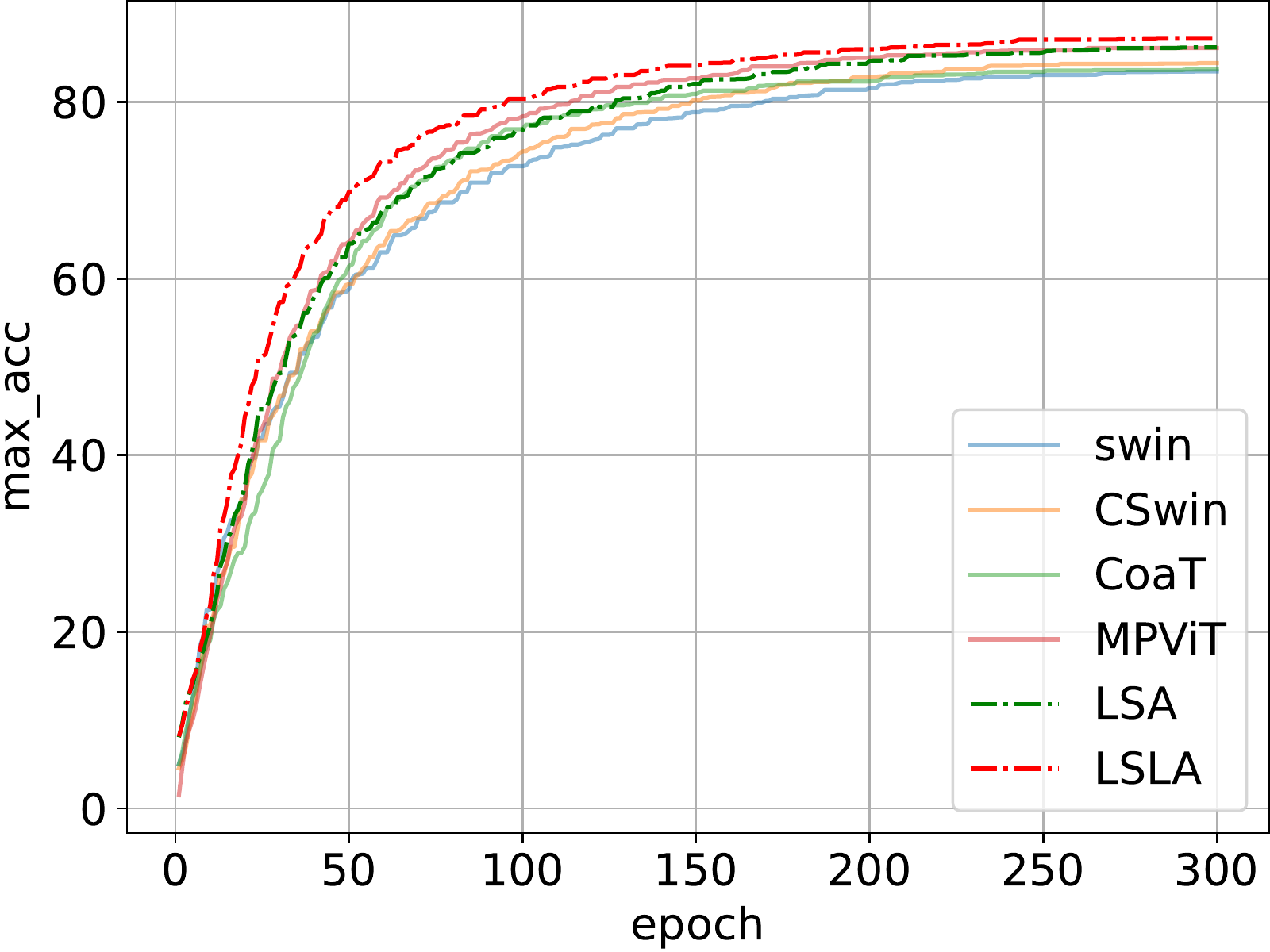}
		\label{fig:train_Mini}
	}
	\caption{ The LSLA is our ViT-LSLA; the LSA is the ViT-LSLA discarding the outer position bias. }
	\label{fig:train}
\end{figure}

Lastly, the LSLA can increase performance and the speed of convergence. In Figure \ref{fig:train}, the convergent speed of LSLA is faster than other counterparts. Note that simply dropping the outer position bias of LSLA, the LSA can result in substantially inferior performance and convergence speed.
\section{Conclusion}
This paper presents a hierarchical vision Transformer with the light self-limited-attention (ViT-LSLA). The LSLA consists of a light self-attention mechanism (LSA) and a self-limited-attention mechanism (SLA).

Firstly, the LSA changes self-attention calculation from Q, K, V to Q, X, X, which can be conveniently applied in current vision Transformers. The LSA can reduce the parameters and computation cost significantly while maintaining performance.

Secondly, this paper introduces the SLA containing a positional information module and a limited-attention module. The former adopts a dynamic scale (DS) and an inner relative position bias to enhance the positional information. The latter uses an outer relative position bias depending on the former to powerfully restrict the large values of attention weight near the query patches. It is beneficial for queries to focus on important and meaningful patches rather than that close and similar ones, which can retain the diversity of information in each query patch.

Finally, the ViT-LSLA demonstrates a competitive performance on several classification datasets and obtains the best efficiency-quality trade-off compared to other counterparts. Therefore, it is hoped that the LSLA will inspire more competitive and novel attention mechanisms.

\bibliography{aaai23}
\end{document}